# Prostate Age Gap (PAG):
## An MRI surrogate marker of aging for prostate cancer detection

Alvaro Fernandez-Quilez[1,2,3,*], Tobias Nordström[3,4], Fredrik Jäderling[5,6], Svein Reidar Kjosavik[7] and Martin Eklund[3]

[1]Department of Computer Science and Electrical Engineering, University of Stavanger, Stavanger.
[2]SMIL, Department of Radiology, Stavanger University Hospital, Stavanger.
[3]Department of Medical Epidemiology and Biostatistics, Karolinska Institutet, Stockholm.
[4]Department of Molecular Medicine and Surgery, Karolinska Institutet, Stockholm.
[5]Department of Radiology, Capio S:t Görän Hospital, Stockholm.
[6]Department of Clinical Sciences, Danderyd Hospital, Karolinska Institutet, Stockholm.
[7]General Practice and Care Coordination Research Group, Stavanger University Hospital, Stavanger.

[*]Corresponding author

Abstract

***Background***: Prostate cancer (PC) MRI-based risk calculators are commonly based on biological (e.g. PSA), MRI markers (e.g. volume), and patient age. Whilst patient age measures the amount of years an individual has existed, *biological age* (BA) might better reflect the physiology of an individual. However, surrogates from prostate MRI and linkage with clinically significant PC (csPC) remain to be explored.

***Purpose***: To obtain and evaluate Prostate Age Gap (PAG) as an MRI marker tool for csPC risk.

***Study type***: Retrospective.

***Population***: A total of 7243 prostate MRI slices from 468 participants who had undergone prostate biopsies. A deep learning model was trained on 3223 MRI slices cropped around the gland from 81 low-grade PC (ncsPC, Gleason score ≤6) and 131 negative cases and tested on the remaining 256 participants.

***Field Strength/Sequence***: 3.0 T axial T2-weighted MRI.

***Assessment***: Chronological age was defined as the age of the participant at the time of the visit and used to train the deep learning model to predict the age of the patient. Following, we obtained PAG, defined as the model predicted age minus the patient's chronological age.

Multivariate logistic regression models were used to estimate the association through odds ratio (OR) and predictive value of PAG and compared against PSA levels and PI-RADS≥ 3.

*Statistical tests*: T-test, Mann-Whitney U test, Permutation test and ROC curve analysis.

*Results*: The multivariate adjusted model showed a significant difference in the odds of clinically significant PC (csPC, Gleason score ≥7) ([OR] =3.78, 95% confidence interval [CI]:2.32-6.16, $p<.001$). PAG showed a better predictive ability when compared to PI-RADS≥ 3 and adjusted by other risk factors, including PSA levels: AUC =0.981 vs AUC =0.704, $p<.001$.

*Data Conclusion*: PAG was significantly associated with the risk of clinically significant PC and outperformed other well-established PC risk factors.

Background

Population-based screening for prostate cancer (PC) has traditionally relied on prostate-specific antigen (PSA) levels in serum for an initial risk assessment of PC [1, 2]. Elevated PSA levels are commonly used to refer patients to a systematic or transrectal ultrasound-guided biopsy and have been credited with a significant decrease in PC mortality [3]. In spite of it, no general consensus exists on the threshold to define abnormal PSA levels, with PSA ≥ 3 ng/mL being a possible choice [3, 4]. Further, its association with over-detection of low grade PC (Gleason Score≤ 6, ncsPC) and subsequent unnecessary testing practices, treatment and related complications has resulted in the absence of governmental bodies recommending organized PSA testing [4, 5].

Over the last decade, multi-parametric MRI (mp-MRI) has emerged as an important component for PC detection, staging, treatment planning and intervention. Some studies have highlighted its benefits in PC screening practices by reducing the number of biopsies and the number of detected ncsPC in the event of elevated PSA levels and a positive MRI [6]. Positive prostate MRIs are defined on the basis of PI-RADS v2.1. scoring system, where PI-RADS≥3 is commonly considered the cut-off to refer a patient to a biopsy, based on a probable presence of a high-grade PC cancer (csPC) [7]. Nevertheless, PI-RADS v.2.1. suffers from limitations in the form of subjective definitions to assign the different categories to detected lesions, with low reported levels of interobserver and intraobserver agreements [7, 8].

A large body of literature has focused on csPC risk calculators able to overcome PI-RADS v2.1. limitations and improve the ability to predict csPC risk by combining the score with demographic and clinical data [7, 9, 10]. Among the data included in the models, prostate volume, PSA, PSA density (PSAd) and chronological age (patient age) are common choices [7, 10]. The latest is a well-established risk factor for csPC, where the incidence increases with advanced age [11, 12]. Whilst chronological age measures the amount of years an individual has existed, *biological age* (BA) has been argued to better reflect physical and mental functions of an individual by incorporating lifestyle, environmental, chronic conditions and genetic factors [13, 14]. That in turn, has been shown to result in a more accurate prediction of age-related diseases including different types of cancers [15, 16, 17].

In this work, we propose to obtain a surrogate of the BA of a patient from their MRI prostate sequence using a deep learning (DL) system. We define the difference between the BA and the chronological patients' age as Prostate Age Gap (PAG) and investigate the association between PAG and csPC risk incidence and its csPC predictive value when adjusted for other common PC risk factors: PSA (ng/mL), prostate volume (mL) and chronological age (years). We *hypothesize* that PAG can serve as a proxy to measure the *biological aging speed process* of the patient and will be associated with an increased risk of csPC independently of other PC risk factors.

## Material and Methods

### Study

PI-CAI (Prostate Imaging: Cancer AI) is a collection of retrospectively collected prostate MRI exams to validate modern AI algorithms and estimate radiologists' csPC detection and diagnosis performance [18]. Patient exams are collected on the basis of suspected csPCa due to elevated PSA levels and, in some cases, abnormal DRE findings. Patients included in the study do not have a history of previous treatment or prior biopsy-confirmed csPC findings.

The main study of PI-CAI consists of 1500 T2-weighted (T2w) and Diffusion Weighted (DW) MRI exams collected from four different centers and two MRI vendors. Out of the provided sequences, 425 cases have biopsy-confirmed csPC, from which 220 are annotated at the pixel level by one trained investigator or one radiology resident, under the supervision of one expert radiologist. Clinical and demographic variables including PSA levels (ng/mL), prostate volume (mL), PSAd ($ng/mL^2$) and patient chronological age (years) are provided together with the sequences when a value was reported during clinical routine [18].

### Data and preprocessing

We use T2-weighted (T2w) axial patient sequences from PI-CAI challenge [18] acquired from 2012 to 2021, together with their paired prostate whole gland masks. Images were acquired with either a Siemens or Philips scanner with an in-plane resolution of 0.5 mm by 0.5 mm and 3 mm slice thickness and with a surface coil. Participants' MRI sequences were included in the study on the basis of *biopsy-confirmed* (systematic, MRI-guided or a combination of both) csPC (Gleason score ≥ 7), ncsPC (Gleason Score ≤ 6) and negative cases. For the reminder of the article and unless otherwise stated, we consider ncsPC to include both biopsy-confirmed low-risk PC and negative examinations.

Patients with missing clinical, demographic data, human annotations or faulty annotations at the time the study was performed were discarded. Further, only baseline visits were considered, where baseline is defined as the first visit of the patient when multiple are available. Figure 1 shows the inclusion criteria of the study. After following the inclusion criteria, 468 participants (354 ncsPC and 114 csPC) T2w MRI sequences were included in the study.

The multi-center and multi-vendor nature of the data used in the study leads to noticeable visual differences in the sequences. To palliate them, we normalize the pixel intensity range and apply a cropping around the prostate gland using the available prostate masks. To avoid edge cases where parts of the prostate gland could be omitted due to restrictive mask ranges, we extend the cropping area 40 pixels in the horizontal and vertical directions. Figure 2 depicts an example of an original sequence and the result after the pre-processing steps.

Deep learning for chronological age prediction

We started by splitting the 354 ncsPC sequences in 60/40% to obtain the train set and independent ncsPC test set, resulting in 212 and 142 sequences, respectively. Cross-contamination is avoided by splitting at the patient level. Table 1 depicts the baseline characteristics of the participants used at training time. At this point, it is important to state that the DL system is trained to *predict, exclusively, the chronological age* of ncsPC patients. Hence, we did not include csPC patients in the training stage of the DL system. The 114 csPC are kept for the independent testing phase, together with the 20% testing sample of 142 ncsPCa. Figure 3 depicts the technical approach to the project, including data splitting.

To maximize the amount of data available at training time, we trained the model at the *slice-level*. That is, we used 2D slices from each patient sequence as inputs of the system. The approach resulted in 3223 ncsPC 2D inputs available during training. We chose a ResNet34 [16] as the DL model based on its simplicity and performance in similar tasks [13]. Rotation and flipping data augmentations were used in an on-line fashion with probability $p = 0.5$ at training time. Details of model selection and performance are detailed in complementary studies and considered to be out of the scope of the present one.

The model is trained with a 5-fold cross-validation (CV) for 120 epochs, where we seek to minimize Mean Absolute Error (MAE) between the predicted age and the chronological age. The resulting prediction of the model is assumed to be a *surrogate of the prostate BA* of the patient, under the assumption that the bigger the residual between the predicted model age and the chronological age the bigger the impact of underlying factors that the model is able to capture. In every fold, the validation loss is closely monitored and the one reaching the minimum is saved and used to report the MAE at the slice level for the fold under consideration. Model training and validation are carried out on an NVIDIA A100-80G GPU.

## Evaluation

We evaluate PAG using the independent test sets for both ncsPC and csPC patients set aside before training the DL system. During evaluation, we calculate the PAG per patient by averaging the PAG for every slice pertaining to a specific patient sequence, giving a unique predicted age per patient. Figure 3 depicts the calculation process followed for every patient.

## Prostate Age Gap (PAG) definition

The prostate age gap (PAG) is defined as the difference between the predicted patient age (surrogate BA) by the DL system and the chronological age of the patient. A positive PAG indicates a large residual between the predicted age of the system and the chronological age, and can be interpreted as the existence of underlying factors present in the MRI sequence and captured by the system that the human naked eye is not able to perceive and that increase the perceived patient age by the system. On the other hand, a negative PAG might be indicative of the presence of underlying factors that positively affect the systems' perception of the patient's age, resulting in a smaller predicted age compared to the chronological one. Figure 3 depicts an example of PAG interpretation.

Statistical analyses

We report continuous variables as mean and standard deviation (mean ± SD), whilst categorical variables are reported as number of occurrences and percentage (N (%)). We perform unpaired T-test, Mann-Whitney U and permutation tests where appropriate to assess the significance of the differences between csPC and ncsPC groups. A standard difference of $p < 0.05$ is considered to be statistically significant. Multivariate logistic regressions are used to estimate the odds ratio (OR) and association between PAG and risk of csPC and in a predictive modeling approach to assess the predictive performance of the proposed marker.

Inference and descriptive analysis

We introduce PAG as a continuous variable and consider different scenarios with a univariate analysis, partially adjusted and fully adjusted models by different well-established risk factors. We start by presenting a descriptive analysis in the test set of the demographic, clinical and PAG for csPC and ncsPC groups. Following, we present Model I unadjusted, Model II adjusted by age (years), volume (mL) and PSA (ng/mL). Model III is adjusted by age (years), volume (mL), PSA (ng/mL) and PI-RADS ≥ 3. Both Model II and Model III are intended to provide a hypothetical *standard* PC screening scenario where PSA and PI-RADS≥3 are considered along with the chronological age of the patient. Model IV and Model V consider a case-scenario where PSA (ng/mL) and volume (mL) in Model II and Model III are replaced with PSAd (ng/mL$^2$). Finally, Model VI considers a scenario where age and PI-RADS ≥ 3 are considered as adjusting factors.

Predictive performance

We use a multivariate logistic regression to estimate the predictive value of PAG (years) adjusted by PSA (ng/mL), prostate volume (mL) and chronological patient age (years) and compare its predictive performance with PI-RADS $\geq$ 3 adjusted by PSA (ng/mL), prostate volume (mL) and chronological patient age (years), simulating common screening PC pathway [18]. Predictive performance is measured in terms of area under the curve (AUC) for the risk of csPC. We obtained an estimate of AUC through bootstrapping of the test set with $n = 1000$ replicates without repetition. Confusion matrices are presented together with AUC for specific FPR rates chosen to emulate common reported PC FPR rates [18]. All analyses are performed in Python 3 with statsmodels 0.14.0 module .

Results

A total of 212 ncsPC participants at baseline (64.78 $\pm$ 6.67 years) were considered for the DL model training. Table 1 depicts the baseline characteristics of the participants. Following, 114 participants with biopsy-confirmed csPC (67.08 $\pm$6.42) and 142 ncsPC (66.86 $\pm$ 7.39) participants that were not included in the DL model training were used in the inference and predictive analysis. As depicted in the baseline characteristics in Table 2, both groups of participants were equally distributed in terms of age. However, csPC patients were more likely to have a higher PSA level, less likely to have a larger prostate and more likely to have a larger PSAd level. As it can also be appreciated in Table 2, PAG was more likely to be substantially larger and positive for csPC patients, indicating that the surrogate BA obtained with DL model was larger when compared to the chronological age. Figure 4 depicts the PAG distributions for csPC and ncsPC patients in the test set.

PAG and risk of csPC

Table 3 shows the different univariate and multivariate logistic regression models adjusted for chronological age (years), PSA (ng/mL), prostate volume (mL), PSAd (ng/mL$^2$) and PI-RADS.

As observed in the univariate model, there was a significant difference in the odds of csPC (OR = 3.07, 95% CI: [2.15, 4.38], $p < .001$).

In the first case-adjusted scenario, we consider a hypothetical case where screening is performed on the basis of PSA (ng/mL) together with the volume of the prostate (mL) and chronological age (years). When partially adjusting the model by them (Model II), there was a significant difference in the adjusted OR per one year increase of PAG of csPC (OR = 3.77, 95% CI: [2.33, 6.11], $p < .001$). Following, we consider a fully adjusted model (Model III) accounting for PI-RADS≥ 3, and find that the significant difference in the adjusted OR of csPC is kept (OR = 3.78, 95% CI: [2.32, 6.16], $p < .001$).

When considering Model IV in which a scenario were PSAd (ng/mL$^2$) is used instead of prostate volume (mL) and PSA (ng/mL), we find that the adjusted OR still remain associated with csPC risk (OR = 3.61, 95% CI: [2.31, 5.64], $p < .001$). Finally, we consider a scenario were only MRI characteristics are considered for screening together with chronological patient age (Model VI) and find an association between the adjusted OR per one year PAG increase and csPC risk (OR = 13.18, 95% CI: [2.88, 60.28], $p < .001$).

Predictive value of PAG for csPC

We compare the predictive value of PAG when adjusted by PSA (ng/mL), chronological patient age (years) and prostate volume (mL) with that of a model including PI-RADS≥ 3 and adjusted for the same factors. Figure 5 depicts the AUC of both models, with a significant improvement of the bootstrapped AUC for the PAG adjusted model when compared to the PI-RADS one (AUC = 0.981, 95% CI: [0.975, 0.987]; AUC = 0.704, 95% CI: [0.652, 0.756], $p < .001$).

Figure 6 depicts the results when examining the AUC under an operating point of False Positive Rate (FPR) of 0.30. We can observe the trade-off of a low FPR at the expense of missing 50% of csPC cases in the case of adjusted PI-RADS≥ 3. In the case of adjusted PAG, we can observe that the model is able to detect all the csPC cases when operating at a 5% FPR. Finally, we analyze the sub-group of patients with PI-RADS≤ 2 but pathologically confirmed csPC. Figure 7 shows the distribution of PAG for the csPC patients with PI-RADS≤ 2 and PI-RADS≥ 3. As it can be observed, the distributions are similar with positive PAG values depicting a larger BA than the chronological patient age.

Discussion

High PSA levels are commonly used as a screening practice for PC, with patients referred to further invasive testing practices such as biopsies on the basis of it. However, PSA is known to be associated with an over-detection of low grade PC (ncsPC) and under-detection of high grade PC (csPC). Alternatives to PSA include using mp-MRI and PI-RADS≥ 3 score as an indication of a probable, but not unequivocal, presence of csPC. Nevertheless, studies have shown the high variability in interobserver and intraobserver rates due to its subjective definitions or MRI quality, leading to the proposal of csPC risk calculators to account for different PC risk factors along with PI-RADS score [7, 8, 9, 10].

The risk of csPC is known to be associated with patient age, where its incidence increases with advanced age [11, 12]. In this retrospective study, we propose to derive from MRI a proxy to measure the *aging speed process of the patient*: the prostate age gap (PAG). Our proposed MRI marker can be considered complementary to commonly used risk factors of csPC such as prostate volume (mL), PI-RADS≥ 3 and common clinical risk factors such as PSA (ng/mL) and PSAd (ng/mL$^2$). In particular, PAG accounts for a proxy of the *biological age*, measured from the T2w axial MRI of a given patient and the chronological age of the patient. That is, PAG accounts for other underlying factors that might be associated with csPC such as lifestyle or environment that are embedded in the *predicted surrogate biological age* by the DL system.

Our retrospective study provides some degree of evidence of the independent association of one year increase in PAG and the odds of csPC when adjusted by PSA (ng/mL), prostate volume (mL) and chronological patient age (years) (OR = 3.77, 95% CI: [2.33, 6.11], $p < .001$). Further, we show the predictive ability of csPC PAG in a multivariate logistic regression when adjusting by the same variables, where we reach an AUC of 0.981 in an independent testing cohort. As far as we know, this is the first study to propose PAG as a risk factor of csPC and a way to calculate the patient's prostate *biological age* directly from MRI, with other studies supporting the idea of biological age and csPC linkage [11, 12, 15, 16]. Our results support that PAG is independently associated with csPC risk even in the presence of PSA and PI-RADS≥ 3, two of the main current measures currently used for PC population-screening practices. Further, PAG shows to reliable predict csPC presence even in the event of PI-RADS≤ 2.

Given the increase in interest and positive results shown by other studies in regards to MRI-based screening [6, 7, 9] as an alternative to PSA or complementary to it [6, 7], we believe that PAG can offer a new and fresh perspective to the screening controversy. Given the need for more reliable MRI-derived risk factors of csPC and new population-based policies for PC that account not only for the patient's age but other factors that better reflect the aging process of the patient (i.e. environmental and lifestyle), PAG offers an alternative that tackles both issues. Moreover, PAG shows to be reliable in the presence of multi-center and multi-vendor data, as reflected in the results obtained in our internal testing cohort used to develop the study [18].

Our study is not the first one to tackle the csPC detection from MRI, with other studies aiming at detecting csPC lesions using deep learning (DL) techniques [7, 20, 21, 22]. In spite of not being directly comparable due to differences in cohorts, our approach shows promising results which surpass those of the ones presented in the literature, even when adjusted by other clinical factors. Moreover, our approach benefits from being easily interpretable as the concept of PAG is intuitive whilst other approaches might not present an alternative that is as intuitive as the one presented in this work, due to the inherent black-box nature of the DL algorithms. Such difficulties can pose ethical and practical challenges in hypothetical scenarios where the tool is deployed in clinical practice, for both the practitioner and the patient perspective [19].

Our findings add new evidence to the limited body of knowledge about the concept of *biological age* and csPC risk. The predicted patient age of the DL system, as a proxy to the biological age of the patient, can be an important measure that highlights the need for more holistic approaches for PC screening where underlying factors such as genetic or environmental ones are considered. Factors like prostate morphology might carry information about the patient that factors such as PSA might not be able to capture. Further, the prostate MRI could be considered as a lens through which the biological age of the body, and, in particular, "prostate physiological" age of the patient can be inferred and allows for a better quantification of the *real aging process* of the patient.

Our study was limited by the sample size and its retrospective nature. Future studies including external evaluations of the DL algorithm are required to verify the validity of our findings and assess its clinical value. The distribution of the population used in the study is not representative of all ethnicities or other important demographic variables, which highlights the need for further external evaluation in the presence of those factors. Moreover, since designing the best DL algorithm for age prediction was considered to be out of the scope of the article, future studies will be required to improve and validate the algorithm validity and to further tune it.

Conclusions

We found that prostate age gap (PAG) derived from T2w MRI images from the prostate was associated with increased risk of clinically significant prostate cancer (csPC). The association was kept after adjusting for clinical and demographic factors such as PSA (ng/mL) levels, prostate volume (mL), patient age (years) and PI-RADS≥ 3. Our new proposed marker has the potential to palliate the overdetection and scoring variability shortcomings in current PC testing practices. Further, it can open an interesting venue related to understanding the aging process of the patient and its association with csPC risk. More studies are required to validate our findings and explore the potential of PAG as a risk factor for prostate cancer.

## List of abbreviations

PC: Prostate Cancer
csPC: Clinically significant prostate cancer
ncsPC: non-clinically significant prostate cancer
mp-MRI: multi-parametric MRI
PAG: Prostate Age Gap
BA: Biological Age
PSA: Prostate Specific Antigen
PSAd: PSA Density
MRI: Magnetic Resonance Imaging
T2w: T2-weighted
T: Tesla

## Declarations

### Ethics approval and consent to participate

All participants provided informed consent in the PI-CAI challenge. The PI-CAI challenge was reviewed and approved by the corresponding ethical committee and consortiums. Since de-identified data was used, requirements of ethical approval were waived. The study was performed in accordance with the Declaration of Helsinki.

### Availability of the data

PI-CAI data is available at the following link:

*https://zenodo.org/record/6517398*

PI-CAI challenge information is available in the following address:

*https://pi-cai.grand-challenge.org/*

PI-CAI prostate annotations are available in the following github repository:

*https://github.com/DIAGNijmegen/picai_labels*

Competing interests

The authors declare that they do not have any competing interests.

References

1. Schröder FH, Hugosson J, Roobol MJ, Tammela TL, Ciatto S, Nelen V, Kwiatkowski M, Lujan M, Lilja H, Zappa M, Denis LJ. Prostate-cancer mortality at 11 years of follow-up. New England Journal of Medicine. 2012 Mar 15;366(11):981-90. doi.org/10.1056/NEJMoa1113135
2. Merriel SW, Pocock L, Gilbert E, Creavin S, Walter FM, Spencer A, Hamilton W. Systematic review and meta-analysis of the diagnostic accuracy of prostate-specific antigen (PSA) for the detection of prostate cancer in symptomatic patients. BMC medicine. 2022 Dec;20(1):1-1. doi.org/10.1186/s12916-021-02230-y
3. Grönberg H, Adolfsson J, Aly M, Nordström T, Wiklund P, Brandberg Y, Thompson J, Wiklund F, Lindberg J, Clements M, Egevad L. Prostate cancer screening in men aged 50–69 years (STHLM3): a prospective population-based diagnostic study. The Lancet oncology. 2015 Dec 1;16(16):1667-76. doi.org/10.1016/S1470-2045(15)00361-7
4. Barry MJ. Screening for prostate cancer--the controversy that refuses to die. New England Journal of Medicine. 2009 Mar 26;360(13):1351. doi.org/10.1056/NEJMe0901166
5. Loeb S, Bjurlin MA, Nicholson J, Tammela TL, Penson DF, Carter HB, Carroll P, Etzioni R. Overdiagnosis and overtreatment of prostate cancer. European urology. 2014 Jun 1;65(6):1046-55. doi.org/10.1016/j.eururo.2013.12.062
6. Eklund M, Jäderling F, Discacciati A, Bergman M, Annerstedt M, Aly M, Glaessgen A, Carlsson S, Grönberg H, Nordström T. MRI-targeted or standard biopsy in prostate

cancer screening. New England journal of medicine. 2021 Sep 2;385(10):908-20. doi.org/10.1056/NEJMoa2100852

7. Turkbey B, Purysko AS. PI-RADS: where next?. Radiology. 2023 Apr 25:223128. doi.org/10.1148/radiol.223128

8. Rosenkrantz AB, Ginocchio LA, Cornfeld D, Froemming AT, Gupta RT, Turkbey B, Westphalen AC, Babb JS, Margolis DJ. Interobserver reproducibility of the PI-RADS version 2 lexicon: a multicenter study of six experienced prostate radiologists. Radiology. 2016 Sep;280(3):793-804. doi.org/10.1148/radiol.2016152542

9. Sedelaar JM, Schalken JA. The need for a personalized approach for prostate cancer management. BMC medicine. 2015 Dec;13:1-3. doi.org/10.1186/s12916-015-0344-1

10. Mehralivand S, Shih JH, Rais-Bahrami S, Oto A, Bednarova S, Nix JW, Thomas JV, Gordetsky JB, Gaur S, Harmon SA, Siddiqui MM. A magnetic resonance imaging–based prediction model for prostate biopsy risk stratification. JAMA oncology. 2018 May 1;4(5):678-85. doi.org/10.1001/jamaoncol.2017.5667

11. Rawla P. Epidemiology of prostate cancer. World journal of oncology. 2019 Apr;10(2):63. doi.org/10.14740/wjon1191

12. Vickers AJ, Sjoberg DD, Ulmert D, Vertosick E, Roobol MJ, Thompson I, Heijnsdijk EA, De Koning H, Atoria-Swartz C, Scardino PT, Lilja H. Empirical estimates of prostate cancer overdiagnosis by age and prostate-specific antigen. BMC medicine. 2014 Dec;12:1-7. doi.org/10.1186/1741-7015-12-26

13. Liem F, Varoquaux G, Kynast J, Beyer F, Masouleh SK, Huntenburg JM, Lampe L, Rahim M, Abraham A, Craddock RC, Riedel-Heller S. Predicting brain-age from multimodal imaging data captures cognitive impairment. Neuroimage. 2017 Mar 1;148:179-88. doi.org/10.1016/j.neuroimage.2016.11.005

14. Bell CG, Lowe R, Adams PD, Baccarelli AA, Beck S, Bell JT, Christensen BC, Gladyshev VN, Heijmans BT, Horvath S, Ideker T. DNA methylation aging clocks: challenges and recommendations. Genome biology. 2019 Dec;20:1-24. doi.org/10.1186/s13059-019-1824-y

15. Fernandez-Quilez A. Deep learning in radiology: ethics of data and on the value of algorithm transparency, interpretability and explainability. AI and Ethics. 2023 Feb;3(1):257-65. doi.org/10.1007/s43681-022-00161-9

16. Zheng Y, Joyce BT, Colicino E, Liu L, Zhang W, Dai Q, Shrubsole MJ, Kibbe WA, Gao T, Zhang Z, Jafari N. Blood epigenetic age may predict cancer incidence and mortality. EBioMedicine. 2016 Mar 1;5:68-73.

17. Dugué PA, Bassett JK, Joo JE, Jung CH, Ming Wong EE, Moreno-Betancur M, Schmidt D, Makalic E, Li S, Severi G, Hodge AM. DNA methylation-based biological aging and cancer risk and survival: Pooled analysis of seven prospective studies. International journal of cancer. 2018 Apr 15;142(8):1611-9.
18. Saha A, Bosma J, Twilt J, van Ginneken B, Yakar D, Elschot M, Veltman J, Fütterer J, de Rooij M. Artificial Intelligence and Radiologists at Prostate Cancer Detection in MRI—The PI-CAI Challenge. In Medical Imaging with Deep Learning, short paper track 2023 Apr 28. doi.org/10.5281/zenodo.6667655
19. A. Fernandez-Quilez, S. V. Larsen, M. Goodwin, T. O. Gulsrud, S. R. Kjosavik and K. Oppedal, "Improving Prostate Whole Gland Segmentation In T2-Weighted MRI With Synthetically Generated Data," 2021 IEEE 18th International Symposium on Biomedical Imaging (ISBI), Nice, France, 2021, pp. 1915-1919, doi: 10.1109/ISBI48211.2021.9433793.
20. Saha A, Hosseinzadeh M, Huisman H. End-to-end prostate cancer detection in bpMRI via 3D CNNs: effects of attention mechanisms, clinical priori and decoupled false positive reduction. Medical image analysis. 2021 Oct 1;73:102155. doi.org/10.1016/j.media.2021.102155
21. A. Fernandez-Quilez, T. Eftestøl, S. R. Kjosavik, M. Goodwin and K. Oppedal, "Contrasting Axial T2W MRI for Prostate Cancer Triage: A Self-Supervised Learning Approach," 2022 IEEE 19th International Symposium on Biomedical Imaging (ISBI), Kolkata, India, 2022, pp. 1-5, doi: 10.1109/ISBI52829.2022.9761573.
22. Schelb P, Kohl S, Radtke JP, Wiesenfarth M, Kickingereder P, Bickelhaupt S, Kuder TA, Stenzinger A, Hohenfellner M, Schlemmer HP, Maier-Hein KH. Classification of cancer at prostate MRI: deep learning versus clinical PI-RADS assessment. Radiology. 2019 Dec;293(3):607-17. doi.org/10.1148/radiol.2019190938

Tables

**Table 1**: Baseline characteristics of ncsPC participants used to train DL model.

**Table 2**: Baseline characteristics of ncsPC and csPC used to evaluate PAG.

**Table 3**: PAG as predictive biomarker of risk of csPC.

**Table 1**: Baseline characteristics of ncsPCa participants used to train DL model.

| *Characteristic* | *ncsPC* <br> *(N = 212)* |
|---|---|
| Age (years) | 64.78 ± 6.67 |
| **PAG (years)** | 2.98 ± 0.03 |
| PSA (ng/mL) | 11.90 ± 17.70 |
| PSA > 3 ng/mL | - |
| *Yes* | 204 (96.22) |
| *No* | 8 (3.78) |
| Prostate volume (mL) | 60.29 ± 30.05 |
| PSAd (ng/mL$^2$) | 0.228 ± 0.442 |
| PI-RADS ≥ 3 | - |
| *Yes* | 146 (68.37) |
| *No* | 66 (31.13) |
| Biopsy Type | - |
| *Systematic* | 66 (31.13) |
| *MRI guided* | 108 (50.94) |
| *MRI (+Systematic)* | 38 (17.93) |

ncsPC = *Gleason score ≤6 and negative cases*

**Table 2**: Baseline characteristics of ncsPC and csPC used to evaluate PAG.

| *Characteristic* | *Group* | | p value |
|---|---|---|---|
| | *ncsPC* (N = 142) | *csPC* (N = 114) | |
| Age (years) | 66.86 ± 7.39 | 67.08 ± 6.42 | 0.802 |
| **PAG (years)** | **-4.11 ± 0.32** | **4.53 ± 2.22** | **< 0.001†** |
| PSA (ng/mL) | 11.91 ± 12.79 | 13.96 ± 10.11 | 0.163 |
| PSA > 3 ng/mL | - | | 0.998 |
| *Yes* | 137 (96.47) | 111 (97.36) | |
| *No* | 5 (3.53) | 3 (2.64) | |
| Prostate volume (mL) | 65.78 ± 38.26 | 54.39 ± 26.60 | **0.007†** |
| PSAd (ng/mL$^2$) | 0.213 ± 0.244 | 0.303 ± 0.239 | **0.003†** |
| PI-RADS ≥ 3 | - | | **0.003†** |
| *Yes* | 96 (67.60) | 96 (84.21) | |
| *No* | 46 (32.39) | 18 (15.78) | |
| Biopsy Type | - | | 0.740 |
| *Systematic* | 46 (32.39) | 18 (15.78) | |
| *MRI guided* | 74 (52.12) | 94 (82.46) | |
| *MRI (+Systematic)* | 22 (15.49) | 2 (1.76) | |

†Statistically significant.

ncsPC = *Gleason score ≤ 6 and negative cases*

csPC = *Gleason score ≥ 7*

20**Table 3**: PAG as predictive biomarker of risk of csPC.

| *Model* | *Adjustment* | *PAG (one year increase)* | | |
|---|---|---|---|---|
| | | OR | 95% CI | p value |
| *I (unadjusted)* | - | 3.07 | [2.15, 4.38] | < 0.001[†] |
| *II (adjusted)* | Age, volume and PSA | 3.77 | [2.33, 6.11] | < 0.001[†] |
| *III (adjusted)* | Age, volume, PSA and PI-RADS | 3.78 | [2.32, 6.16] | < 0.001[†] |
| *IV (adjusted)* | Age and PSAd | 3.61 | [2.31, 5.64] | < 0.001[†] |
| *V (adjusted)* | Age, PSAd and PI-RADS | 13.18 | [2.88, 60.28] | < 0.001[†] |
| *VI (adjusted)* | Age and PI-RADS | 3.15 | [2.17, 4.58] | < 0.001[†] |

[†]Statistically significant.

Figures

**Figure 1.** Patient inclusion criteria.

**Figure 2**. Pre-processing including normalization and prostate gland cropping.

**Figure 3.** Technical approach to the project.

**Figure 4.** Prostate Age Gap (PAG) distributions for the ncsPC and csPC patients in the test set.

**Figure 5.** Multivariate logistic regression and predictive ability of PAG when adjusted by other risk factors.

**Figure 6.** Confusion matrices for multivariate logistic regression and PAG when adjusted (FPR @ 0.10), PI-RADS adjusted (FPR @ 0.30) and PI-RADS unadjusted (FPR @ 0.60).

**Figure 7.** Prostate Age Gap (PAG) distribution for PI-RADS ≤ 2 and csPC patients.

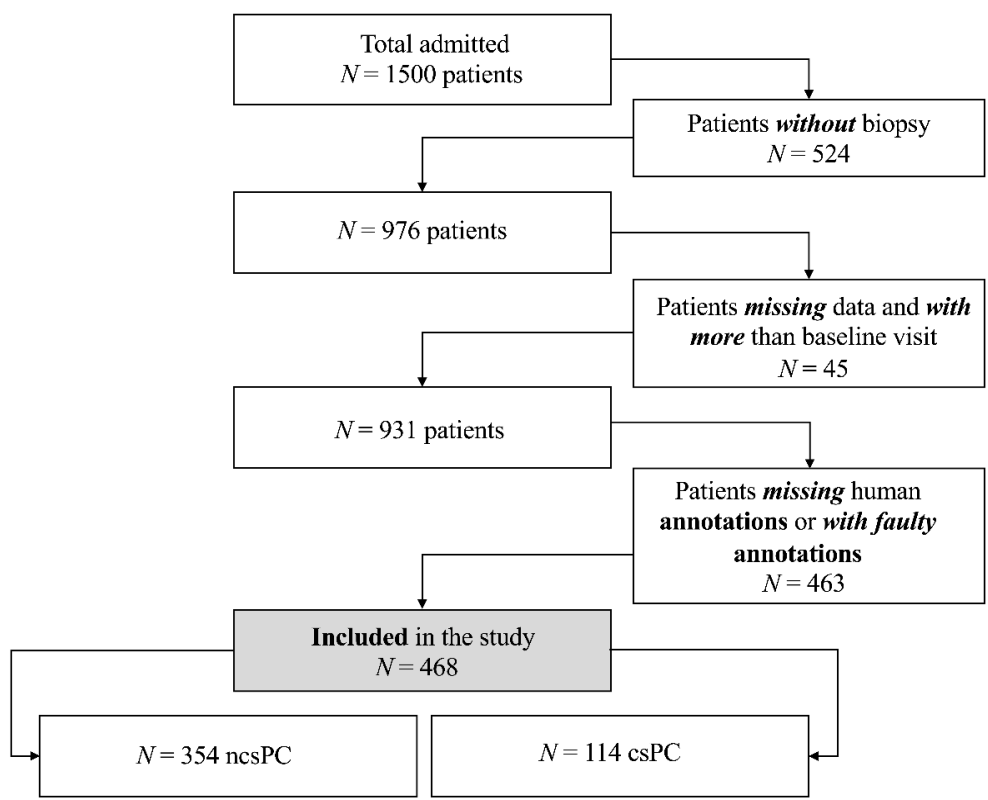

**Figure 1.** Patient inclusion criteria.

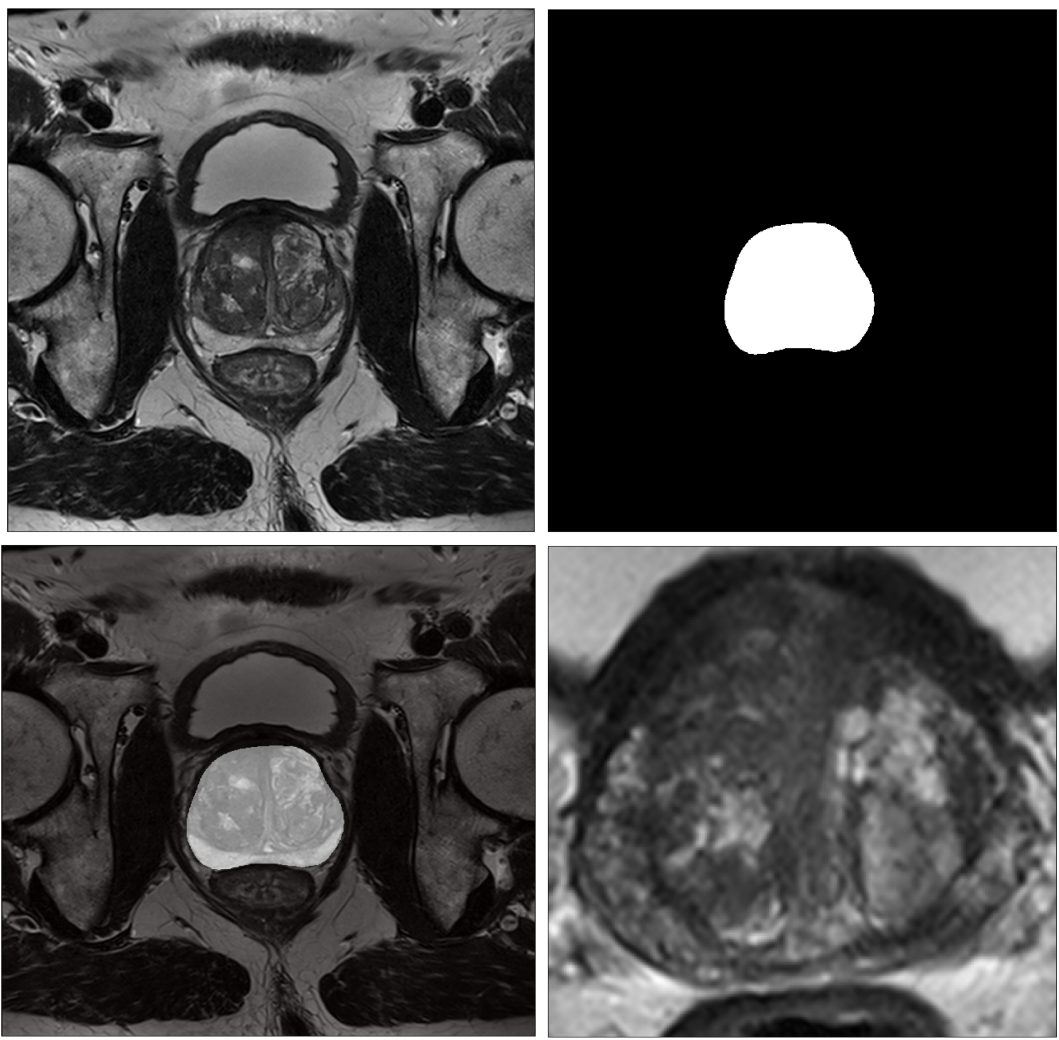

**Figure 2**. Pre-processing including normalization and prostate gland cropping.

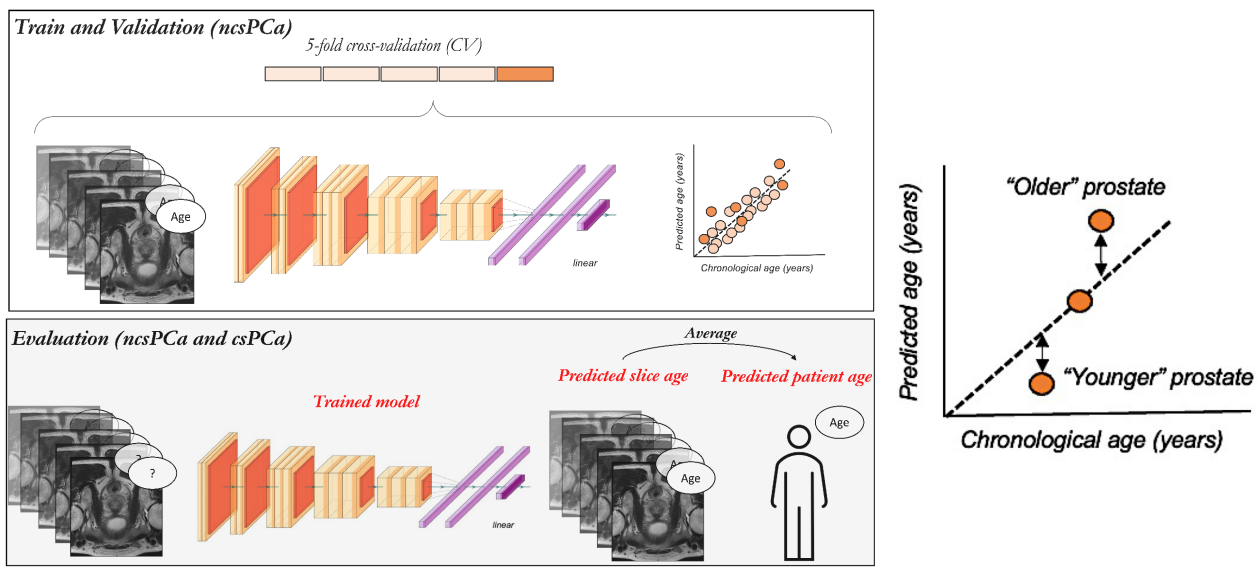

**Figure 3.** Technical approach to the project.

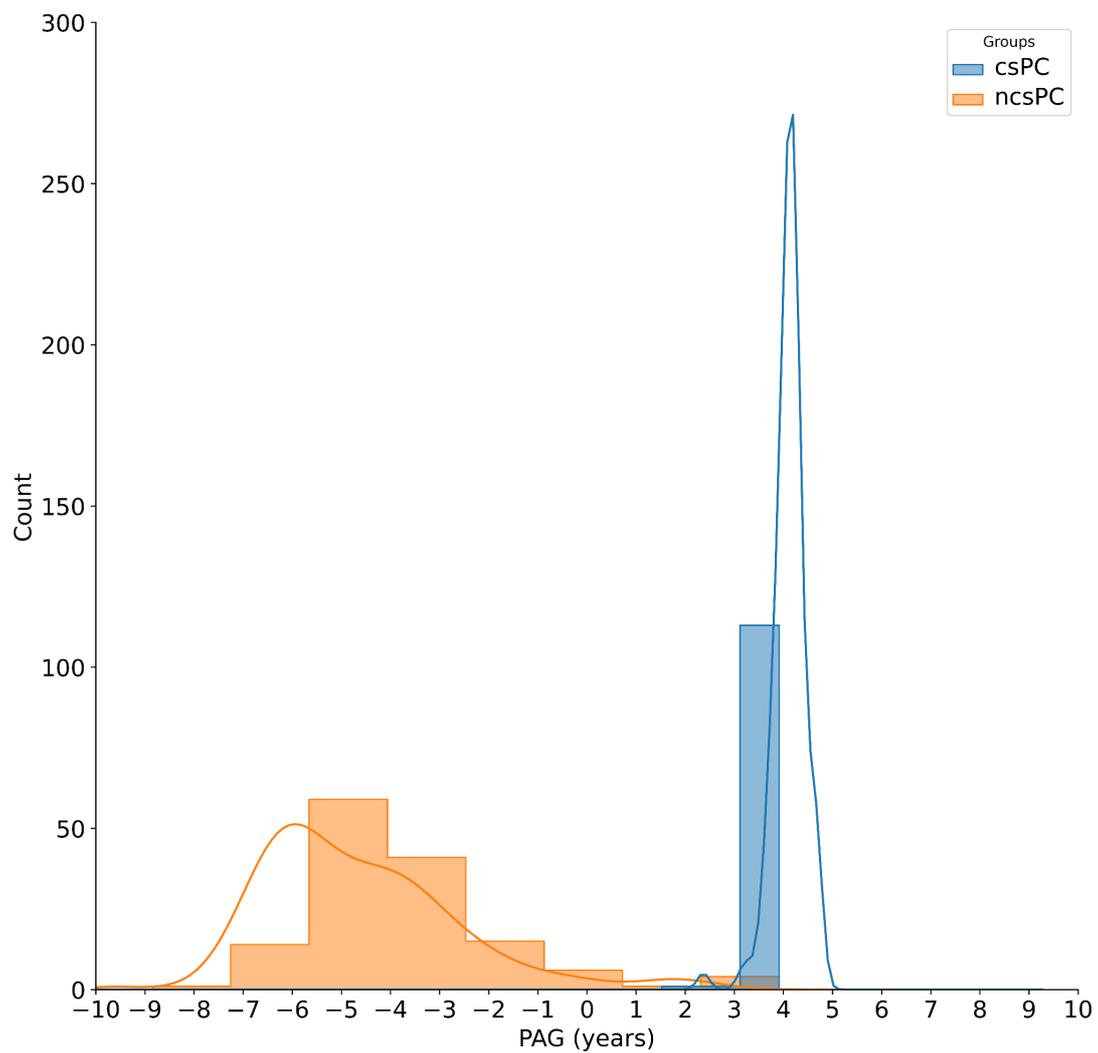

**Figure 4.** Prostate Age Gap (PAG) distributions for the ncsPC and csPC patients in the test set.

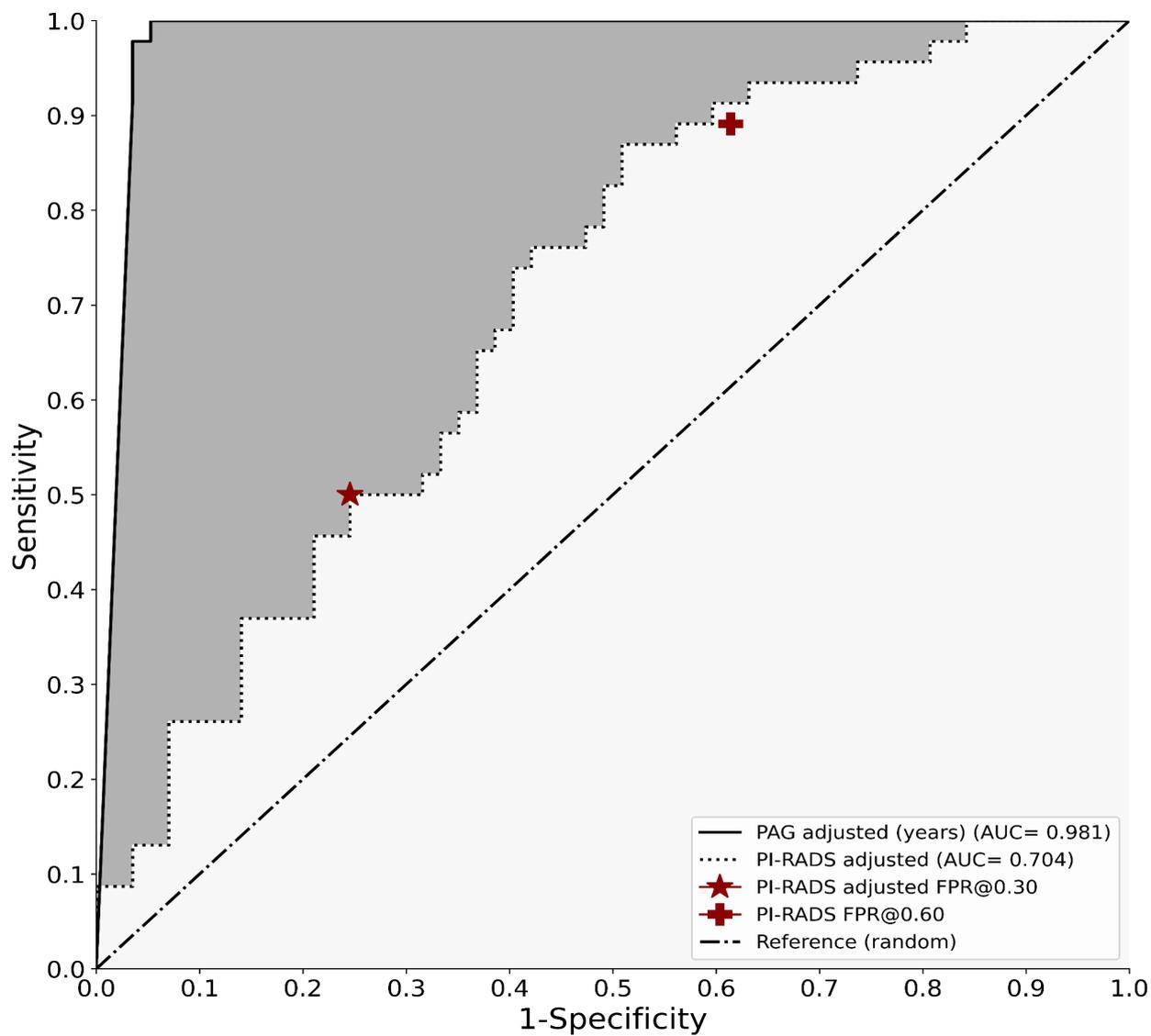

**Figure 5.** Multivariate logistic regression and predictive ability of PAG when adjusted by other risk factors.

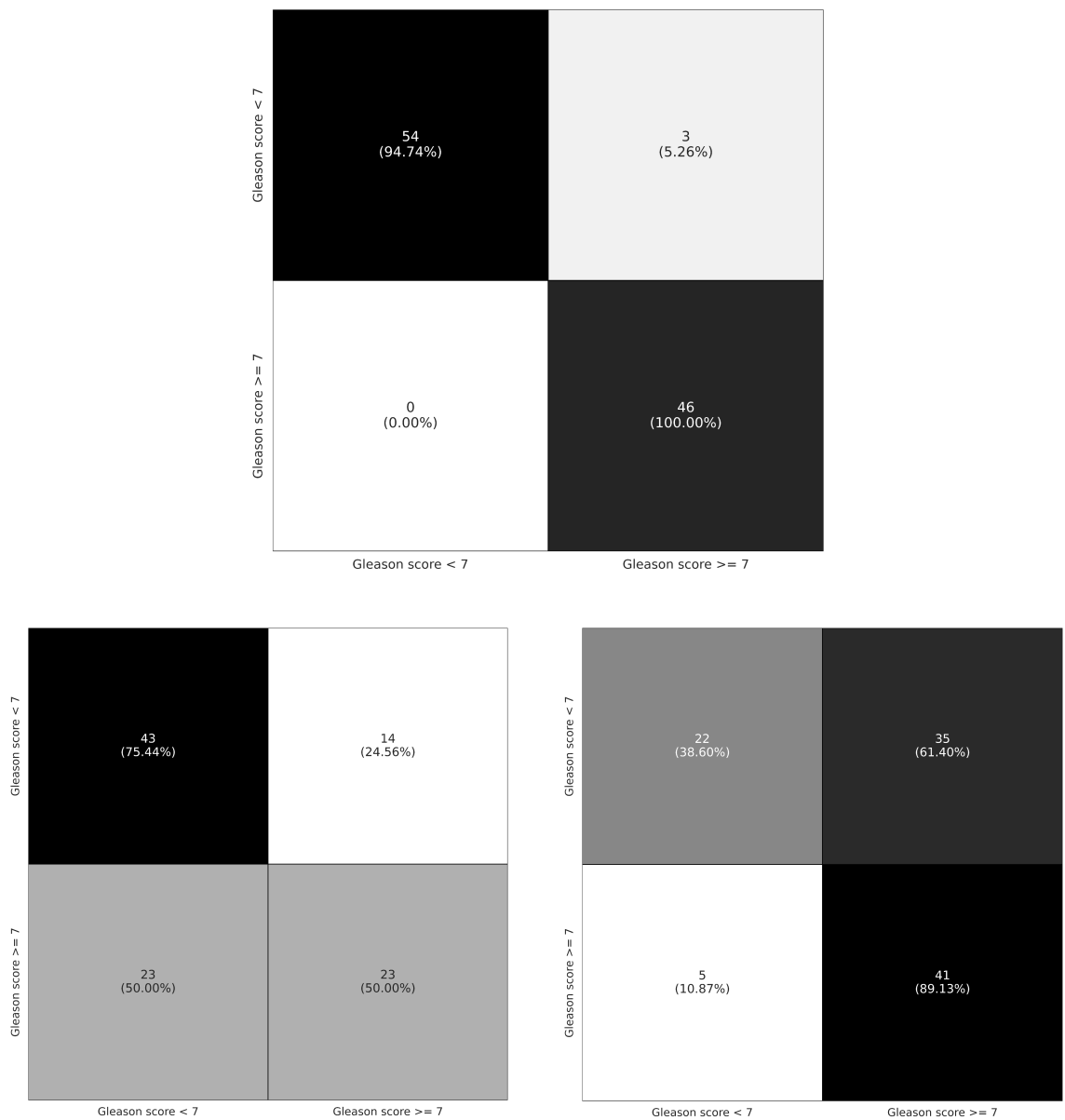

**Figure 6.** Confusion matrices for multivariate logistic regression and PAG when adjusted (FPR @ 0.10), PI-RADS adjusted (FPR @ 0.30) and PI-RADS unadjusted (FPR @ 0.60).

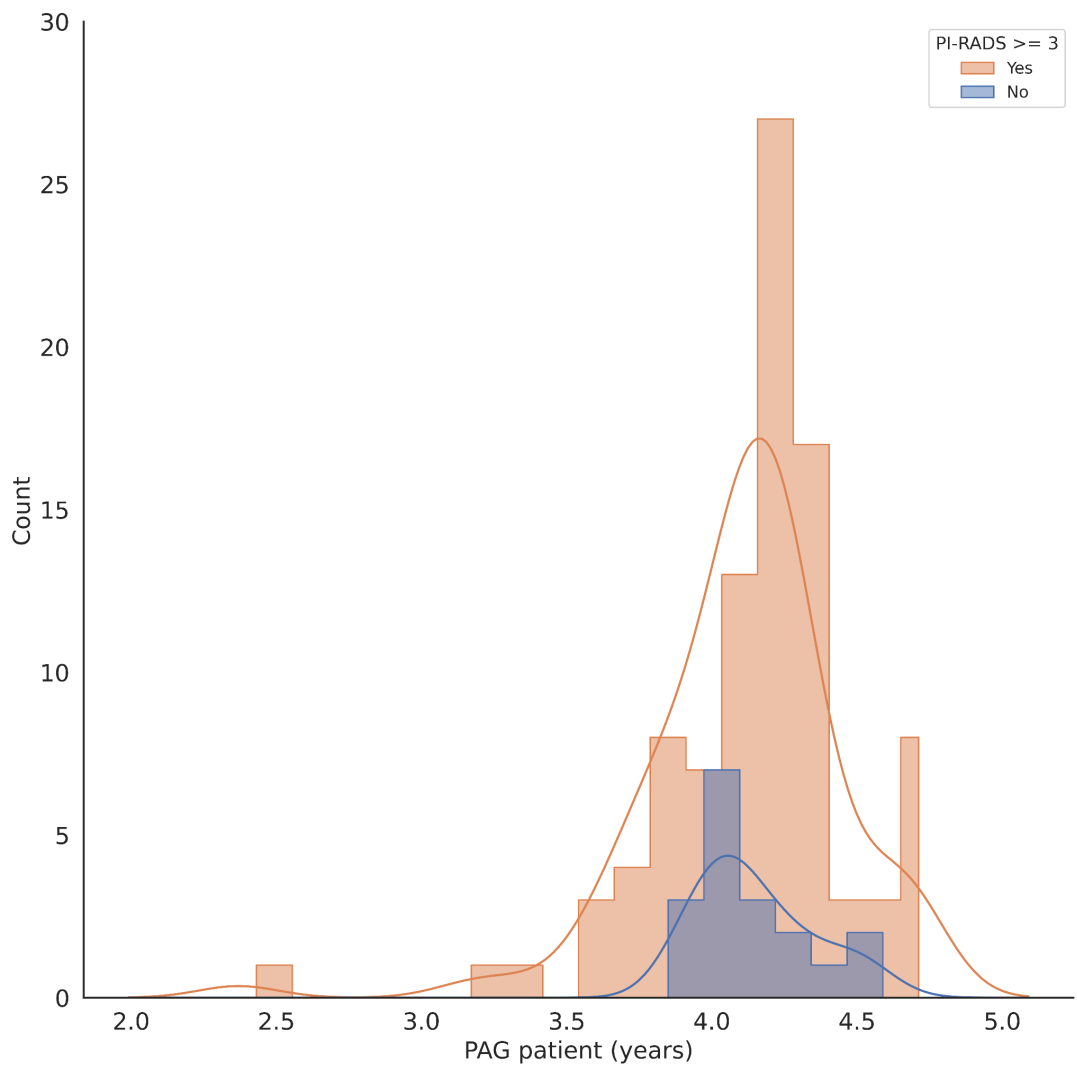

**Figure 7.** Prostate Age Gap (PAG) distribution for PI-RADS ≤ 2 and csPC patients.